\pdfoutput=1

\documentclass[11pt]{article}

\usepackage[]{emnlp2021}

\usepackage{times}
\usepackage{latexsym}
\usepackage{graphicx}
\usepackage{amssymb}
\usepackage{amsmath}
\usepackage{multirow}
\usepackage{array}
\usepackage{caption}
\usepackage[labelformat=simple]{subcaption}
\usepackage{footnote}
\usepackage{enumitem}

\usepackage[T1]{fontenc}

\usepackage[utf8]{inputenc}

\usepackage{microtype}

\usepackage{booktabs}

\makesavenoteenv{caption}
\makesavenoteenv{figure}
\newcolumntype{H}{>{\setbox0=\hbox\bgroup}c<{\egroup}@{}}

\newcommand{\comment}[1]{}

\newcommand*{\inlinegraphics}[1]{%
    \raisebox{0.0\baselineskip}{%
        \includegraphics[
        height=\baselineskip,
        width=\baselineskip,
        keepaspectratio,
        ]{#1}%
    }%
}

\title{The Power of Scale for Parameter-Efficient Prompt Tuning}

\author{
Brian Lester$^*$ \enskip Rami Al-Rfou \enskip Noah Constant \\
Google Research \\
\texttt{\{brianlester,rmyeid,nconstant\}@google.com} \\
}

\begin{document}
\maketitle
\begin{abstract}
In this work, we explore ``prompt tuning,'' a simple yet effective mechanism for learning ``soft prompts'' to condition frozen language models to perform specific downstream tasks.
Unlike the discrete text prompts used by \mbox{GPT-3}, soft prompts are learned through backpropagation and can be tuned to incorporate signals from any number of labeled examples.
Our end-to-end learned approach outperforms \mbox{GPT-3's} few-shot learning by a large margin.
More remarkably, through ablations on model size using T5, we show that prompt tuning becomes more competitive with scale: as models exceed billions of parameters, our method ``closes the gap'' and matches the strong performance of model tuning (where all model weights are tuned).
This finding is especially relevant because large models are costly to share and serve and the ability to reuse one frozen model for multiple downstream tasks can ease this burden.
Our method can be seen as a simplification of the recently proposed ``prefix tuning'' of \citet{li_2021_prefix_tuning} and we provide a comparison to this and other similar approaches.
Finally, we show that conditioning a frozen model with soft prompts confers benefits in robustness to domain transfer and enables efficient ``prompt ensembling.''
\end{abstract}

\renewcommand{\thefootnote}{$^*$}
\footnotetext[1]{Work done as a Google AI Resident.}
\renewcommand\thefootnote{\arabic{footnote}}

\section{Introduction}

With the wide success of pre-trained large language models, a range of techniques has arisen to adapt these general-purpose models to downstream tasks.
ELMo \cite{peters-etal-2018-deep} proposed freezing the pre-trained model and learning a task-specific weighting of its per-layer representations. However, since GPT \cite{gpt} and BERT \cite{devlin-etal-2019-bert}, the dominant adaptation technique has been \textbf{model tuning} (or ``fine-tuning''), where all model parameters are tuned during adaptation, as proposed by \citet{howard-ruder-2018-universal}.

\begin{figure}[h!]
    \centering
    \includegraphics[width=0.9\columnwidth]{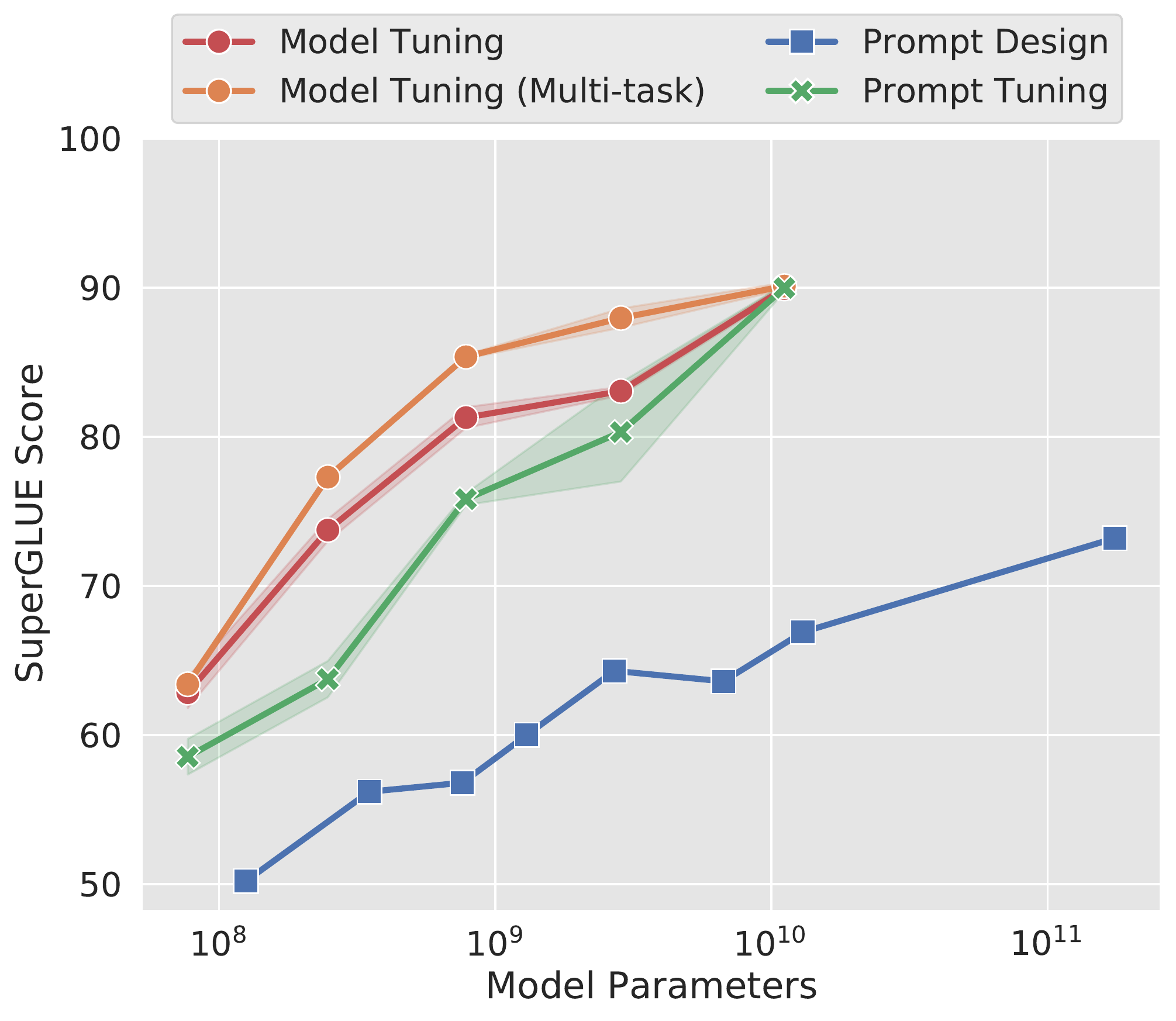}\hspace{1.5ex}
    \caption{Standard \textbf{model tuning} of T5 achieves strong performance, but requires storing separate copies of the model for each end task. Our \textbf{prompt tuning} of T5 matches the quality of model tuning as size increases, while enabling the reuse of a single frozen model for all tasks. Our approach significantly outperforms few-shot \textbf{prompt design} using \mbox{GPT-3}. We show mean and standard deviation across $3$ runs for tuning methods.} 
    \label{fig:model-size}
\end{figure}

More recently, \citet{brown_2020_gpt3} showed that \textbf{prompt design} (or ``priming'') is surprisingly effective at modulating a frozen \mbox{GPT-3} model's behavior through text prompts.
Prompts are typically composed of a task description and/or several canonical examples. This return to ``freezing'' pre-trained models is appealing, especially as model size continues to increase. Rather than requiring a separate copy of the model for each downstream task, a single generalist model can simultaneously serve many different tasks.

\begin{figure}[h!]
    \centering
    \includegraphics[width=\columnwidth]{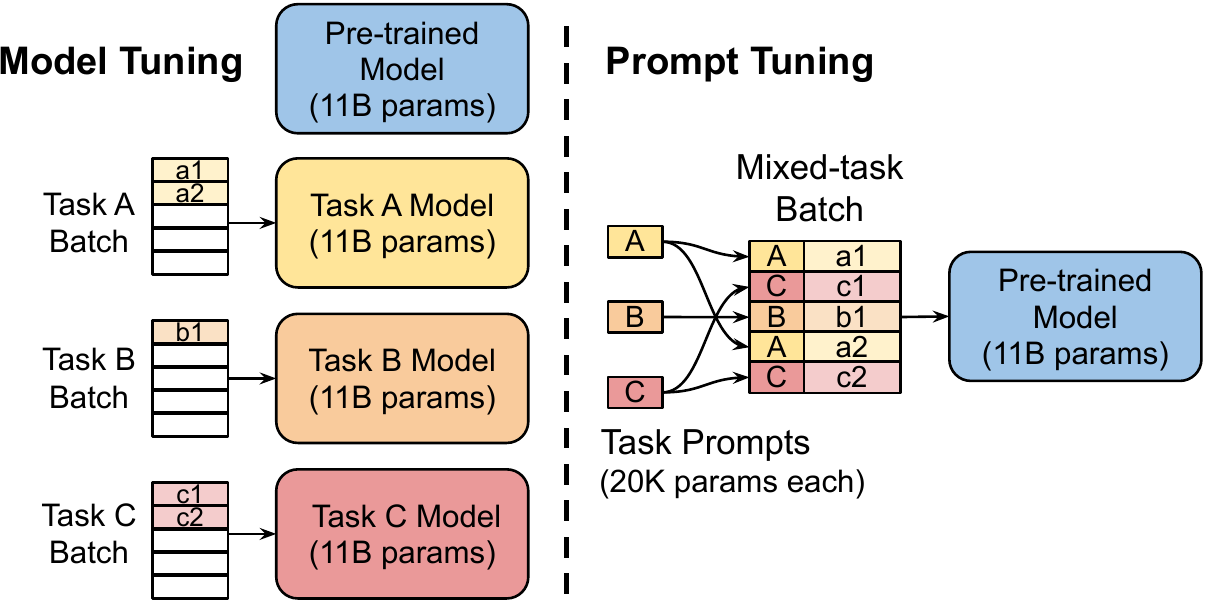}
    \caption{\textbf{Model tuning} requires making a task-specific copy of the entire pre-trained model for each downstream task and inference must be performed in separate batches. \textbf{Prompt tuning} only requires storing a small task-specific prompt for each task, and enables mixed-task inference using the original pre-trained model. With a T5 ``XXL'' model, each copy of the tuned model requires $11$ billion parameters. By contrast, our tuned prompts would only require $20{,}480$ parameters per task---a reduction of \emph{over five orders of magnitude}---assuming a prompt length of $5$ tokens.} 
    \label{fig:diagram}
\end{figure}

Unfortunately, prompt-based adaptation has several key drawbacks.
Task description is error-prone and requires human involvement, and the effectiveness of a prompt is limited by how much conditioning text can fit into the model's input.
As a result, downstream task quality still lags far behind that of tuned models.
For instance, \mbox{GPT-3} 175B few-shot performance on SuperGLUE is $17.5$ points below fine-tuned T5-XXL \cite{raffel_2020_t5} ($71.8$ vs.~$89.3$) despite using $16$ times more parameters.

Several efforts to automate prompt design have been recently proposed.
\citet{shin-etal-2020-autoprompt} propose a search algorithm over the discrete space of words, guided by the downstream application training data.
While this technique outperforms manual prompt design, there is still a gap relative to model tuning.

\citet{li_2021_prefix_tuning} propose ``prefix tuning'' and show strong results on generative tasks. This method freezes the model parameters and backpropagates the error during tuning to prefix activations prepended to each layer in the encoder stack, including the input layer.
\citet{hambardzumyan_2021_warp} simplify this recipe by restricting the trainable parameters to the input and output sub-networks of a masked language model, and show reasonable results on classifications tasks.

In this paper, we propose \textbf{prompt tuning} as a further simplification for adapting language models. We freeze the entire pre-trained model and only allow an additional $k$ tunable tokens per downstream task to be prepended to the input text. This ``soft prompt'' is trained end-to-end and can condense the signal from a full labeled dataset, allowing our method to outperform few-shot prompts and close the quality gap with model tuning (Figure~\ref{fig:model-size}). At the same time, since a single pre-trained model is recycled for all downstream tasks, we retain the efficient serving benefits of frozen models (Figure~\ref{fig:diagram}).

While we developed our method concurrently with \citet{li_2021_prefix_tuning} and \citet{hambardzumyan_2021_warp}, we are the first to show that prompt tuning alone (with no intermediate-layer prefixes or task-specific output layers) is sufficient to be competitive with model tuning.
Through detailed experiments in sections~\ref{sec:tuning}--\ref{sec:results}, we demonstrate that language model capacity is a key ingredient for these approaches to succeed. As Figure~\ref{fig:model-size} shows, \emph{prompt tuning becomes more competitive with scale.}

We compare with similar approaches in Section~\ref{sec:previous_work}. Explicitly separating task-specific parameters from the ``generalist'' parameters needed for general language-understanding has a range of additional benefits. We show in Section~\ref{sec:shift} that by capturing the task definition in the prompt while keeping the generalist parameters fixed, we are able to achieve better resilience to domain shifts. In Section~\ref{sec:ensemble}, we show that ``prompt ensembling'', learning multiple prompts for the same task, can boost quality and is more efficient than classic model ensembling. Finally, in Section~\ref{sec:interpretability}, we investigate the interpretability of our learned soft prompts. In sum, our key contributions are:

\begin{enumerate} [topsep=3pt,itemsep=-1ex,partopsep=1ex,parsep=1ex]
\item Proposing prompt tuning and showing its competitiveness with model tuning in the regime of large language models.
\item Ablating many design choices, and showing quality and robustness improve with scale.
\item Showing prompt tuning outperforms model tuning on domain shift problems.
\item Proposing ``prompt ensembling'' and showing its effectiveness.
\end{enumerate}

\section{Prompt Tuning}
\label{sec:tuning}

Following the ``text-to-text'' approach of T5 \cite{raffel_2020_t5}, we cast all tasks as text generation. Instead of modeling classification as the probability of an output class given some input, $\Pr(y|X)$, where $X$ is a series of tokens and $y$ is a single class label, we now model it as conditional generation, where $Y$ is a sequence of tokens that represent a class label. T5 models classification as $\Pr_{\theta}(Y | X)$, parameterized by the weights, $\theta$, of the transformers \cite{vaswani2017attention} that make up its encoder and decoder.

Prompting is the approach of adding extra information for the model to condition on during its generation of $Y$. Normally, prompting is done by prepending a series of tokens, $P$, to the input $X$, such that the model maximizes the likelihood of the correct $Y$, $\Pr_{\theta}(Y|[P;X])$, while keeping the model parameters, $\theta$, fixed. In \mbox{GPT-3}, the representations of the prompt tokens, $P = \{p_1, p_2, \dots, p_n\}$, are part of the model's embedding table, parameterized by the frozen $\theta$. Finding an optimal prompt thus requires the selection of prompt tokens, through either manual search or non-differentiable search methods \cite{jiang-etal-2020-know,shin-etal-2020-autoprompt}. Prompt tuning removes the restriction that the prompt $P$ be parameterized by $\theta$; instead the prompt has its own dedicated parameters, $\theta_P$, that can be updated. While prompt \emph{design} involves selecting prompt tokens from a fixed vocabulary of frozen embeddings, prompt \emph{tuning} can be thought of as using a fixed prompt of special tokens, where only the embeddings of these prompt tokens can be updated. Our new conditional generation is now $\Pr_{\theta;\theta_P}(Y | [P;X])$ and can be trained by maximizing the likelihood of $Y$ via backpropagation, while only applying gradient updates to $\theta_P$.

Given a series of $n$ tokens, $\{x_1, x_2, \dots, x_n\}$, the first thing T5 does is embed the tokens, forming a matrix $X_e \in \mathbb{R}^{n \times e}$ where $e$ is the dimension of the embedding space. Our soft-prompts are represented as a parameter $P_e \in \mathbb{R}^{p \times e}$, where $p$ is the length of the prompt. Our prompt is then concatenated to the embedded input forming a single matrix $[P_e;X_e] \in \mathbb{R}^{(p+n)\times e}$ which then flows though the encoder-decoder as normal. Our models are trained to maximize the probability of $Y$, but only the prompt parameters $P_e$ are updated.

\subsection{Design Decisions}

There are many possible ways to initialize the prompt representations. The simplest is to train from scratch, using random initialization. A more sophisticated option is to initialize each prompt token to an embedding drawn from the model's vocabulary. Conceptually, our soft-prompt modulates the frozen network's behavior in the same way as text preceding the input, so it follows that a word-like representation might serve as a good initialization spot. For classification tasks, a third option is to initialize the prompt with embeddings that enumerate the output classes, similar to the ``verbalizers'' of \citet{schick-schutze-2021-exploiting}. Since we want the model to produce these tokens in the output, initializing the prompt with the embeddings of the valid target tokens should prime the model to restrict its output to the legal output classes. 

Another design consideration is the length of the prompt. The parameter cost of our method is $EP$, where $E$ is the token embedding dimension and $P$ is the prompt length. The shorter the prompt, the fewer new parameters must be tuned, so we aim to find a minimal length that still performs well.

\subsection{Unlearning Span Corruption}
\label{sec:span_corruption}

Unlike autoregressive language models like \mbox{GPT-3}, the T5 models we experiment with use an encoder-decoder architecture and pre-train on a span corruption objective. Specifically, T5 is tasked with ``reconstructing'' masked spans in the input text, which are marked with unique sentinel tokens. The target output text consists of all the masked content, separated by sentinels, plus a final sentinel. For instance, from the text ``Thank you for inviting me to your party last week'' we might construct a pre-training example where the input is ``Thank you $\langle$X$\rangle$ me to your party $\langle$Y$\rangle$ week'' and the target output is ``$\langle$X$\rangle$ for inviting $\langle$Y$\rangle$ last $\langle$Z$\rangle$''.

While \citet{raffel_2020_t5} find this architecture and pre-training objective more effective than traditional language modeling, we hypothesize that this setup is not a good fit for producing a frozen model that can be readily controlled through prompt tuning. In particular, a T5 model pre-trained exclusively on span corruption, such as T5.1.1, has never seen truly natural input text (free of sentinel tokens), nor has it ever been asked to predict truly natural targets. In fact, due to the details of T5's span corruption preprocessing, every pre-training target will begin with a sentinel. While this ``unnatural'' tendency to output sentinels is easy to overcome through fine-tuning, we suspect that it would be much harder to override through a prompt alone, as the decoder priors cannot be adjusted.

Given these concerns, we experiment with T5 models in three settings. (1) ``Span Corruption'': We use pre-trained T5 off-the-shelf as our frozen model, and test its ability to output the expected text for downstream tasks. (2) ``Span Corruption + Sentinel'': We use the same model, but prepend all downstream targets with a sentinel, so as to more closely resemble the targets seen in pre-training. (3) ``LM Adaptation'': We continue T5's self-supervised training for a small number of additional steps, but using the ``LM'' objective discussed by \citet{raffel_2020_t5}; given a natural text prefix as input, the model must produce the natural text continuation as output. Crucially, this adaptation happens \emph{only once}, producing a single frozen model that we can reuse for prompt tuning across any number of downstream tasks.

Through LM adaptation, we hope to ``quickly'' transform T5 into a model more similar to \mbox{GPT-3}, which always outputs realistic text, and is known to respond well to prompts as a ``few-shot learner''. It is not obvious how successful this late-stage transformation will be compared to pre-training from scratch, and it has not been investigated previously to our knowledge. As such, we experiment with various lengths of adaptation up to 100K steps.

\begin{figure*}[ht!]
\centering

\begin{subfigure}[b]{\columnwidth}
     \centering
     \includegraphics[width=0.75\columnwidth, trim=5 10 15 5, clip]{
     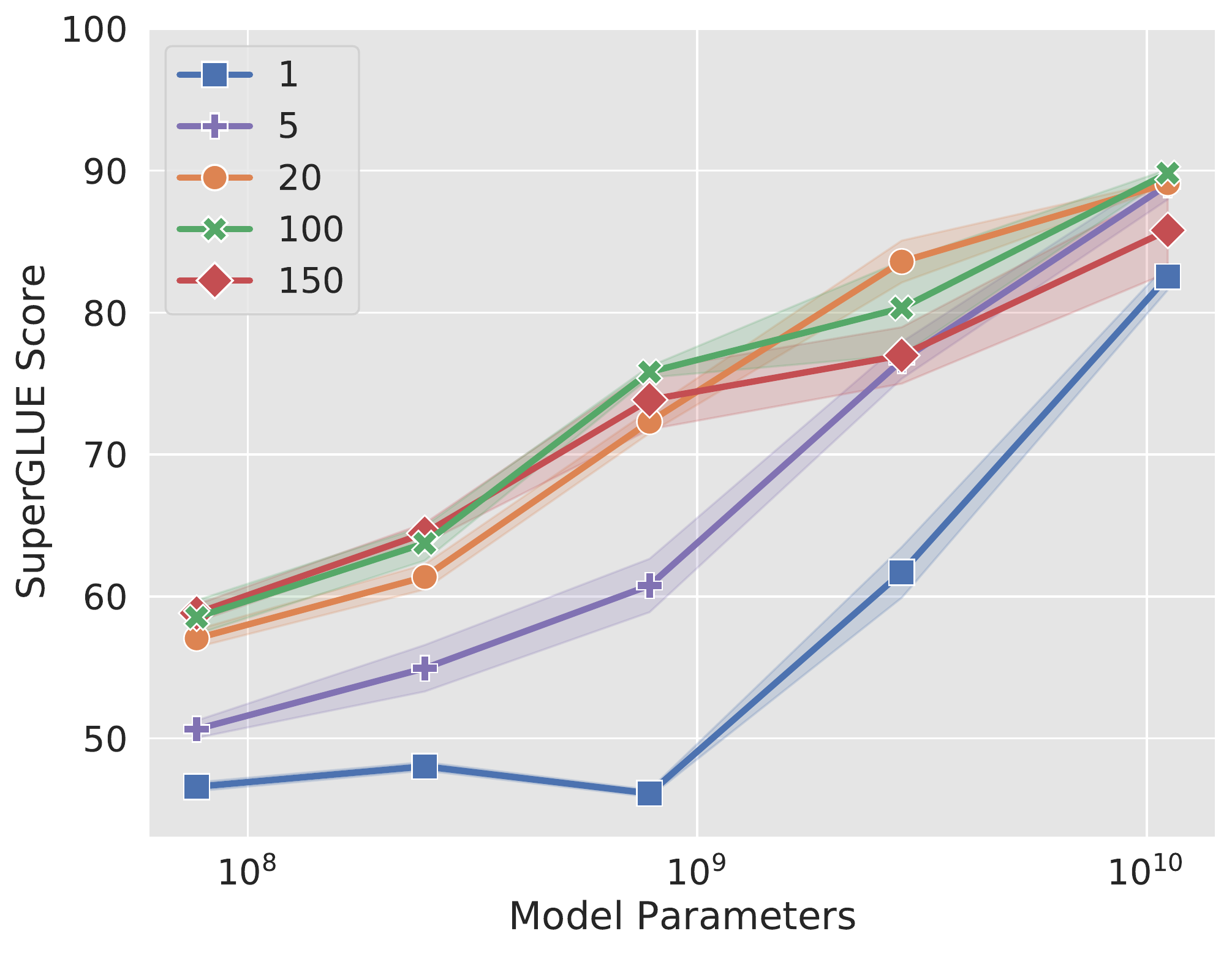}
     \caption{Prompt length}
     \label{fig:ablate-length}
 \end{subfigure}
\begin{subfigure}[b]{\columnwidth}
     \centering
     \includegraphics[width=0.75\columnwidth, trim=5 10 15 5, clip]{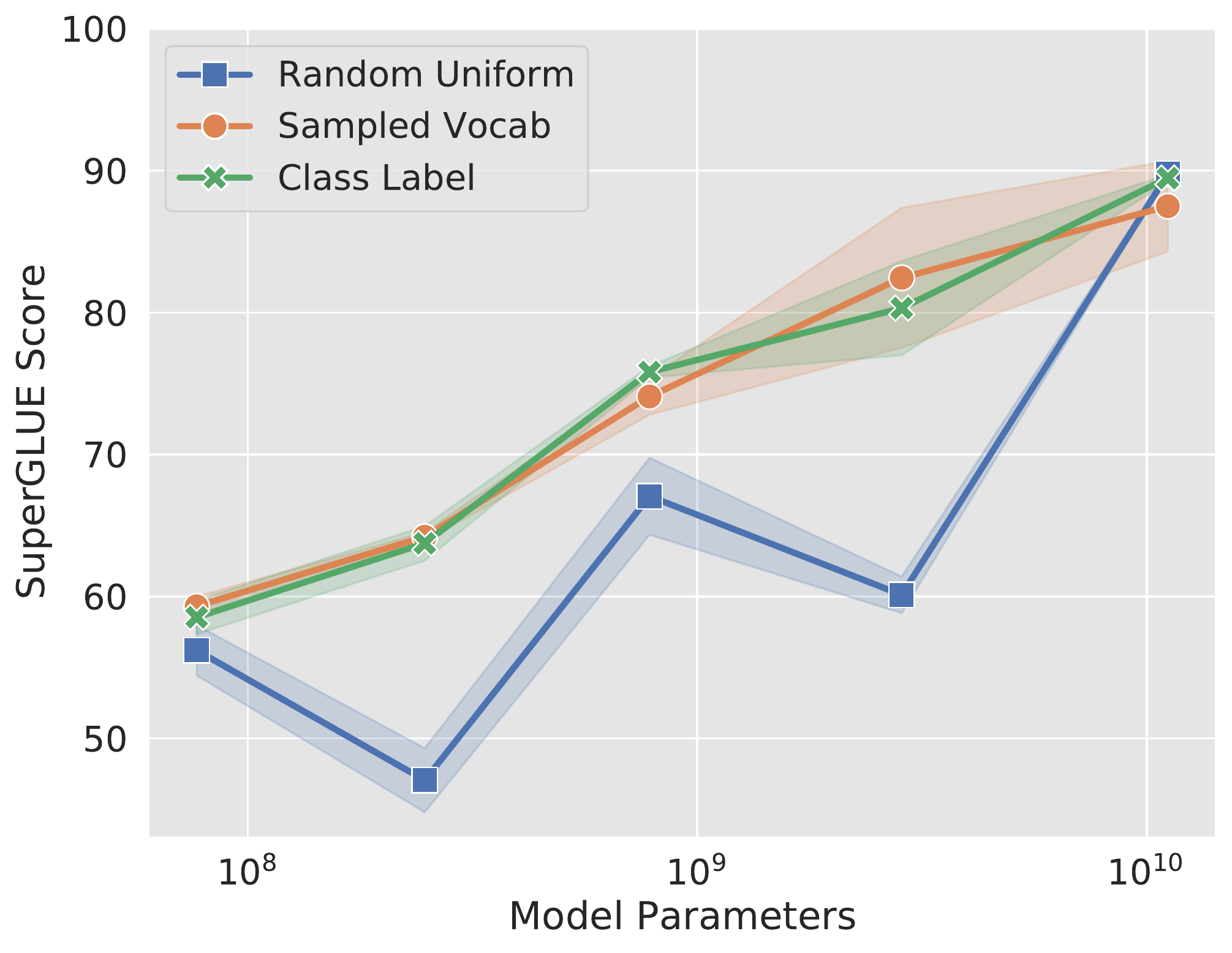}
     \caption{Prompt initialization}
     \label{fig:ablate-init}
\end{subfigure}

\vspace{1ex}

\begin{subfigure}[b]{\columnwidth}
     \centering
     \includegraphics[width=0.75\columnwidth, trim=5 10 15 5, clip]{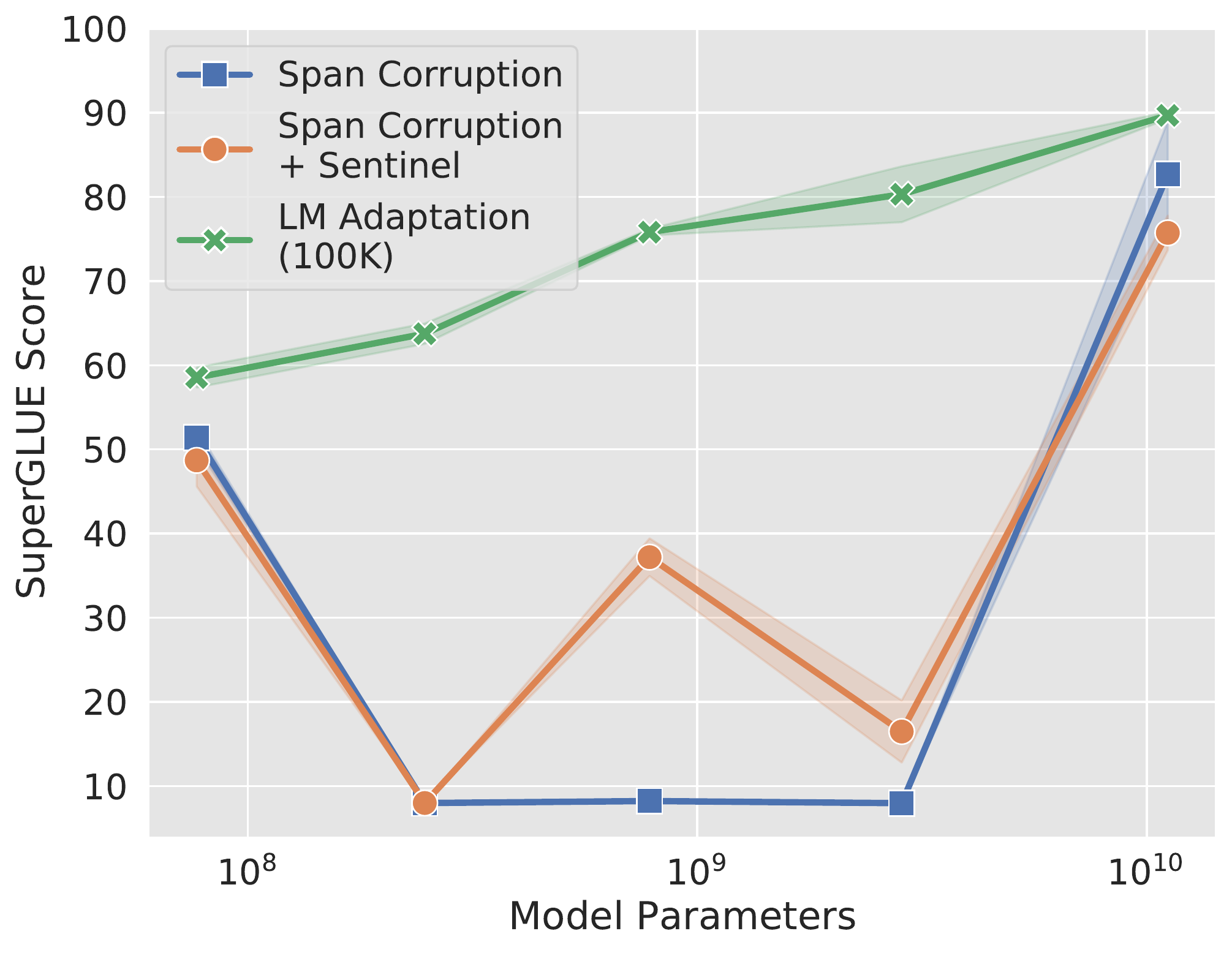}
     \caption{Pre-training method}
     \label{fig:ablate-pretrain}
\end{subfigure}     
\begin{subfigure}[b]{\columnwidth}
     \centering
     \includegraphics[width=0.75\columnwidth, trim=5 10 15 5, clip]{
     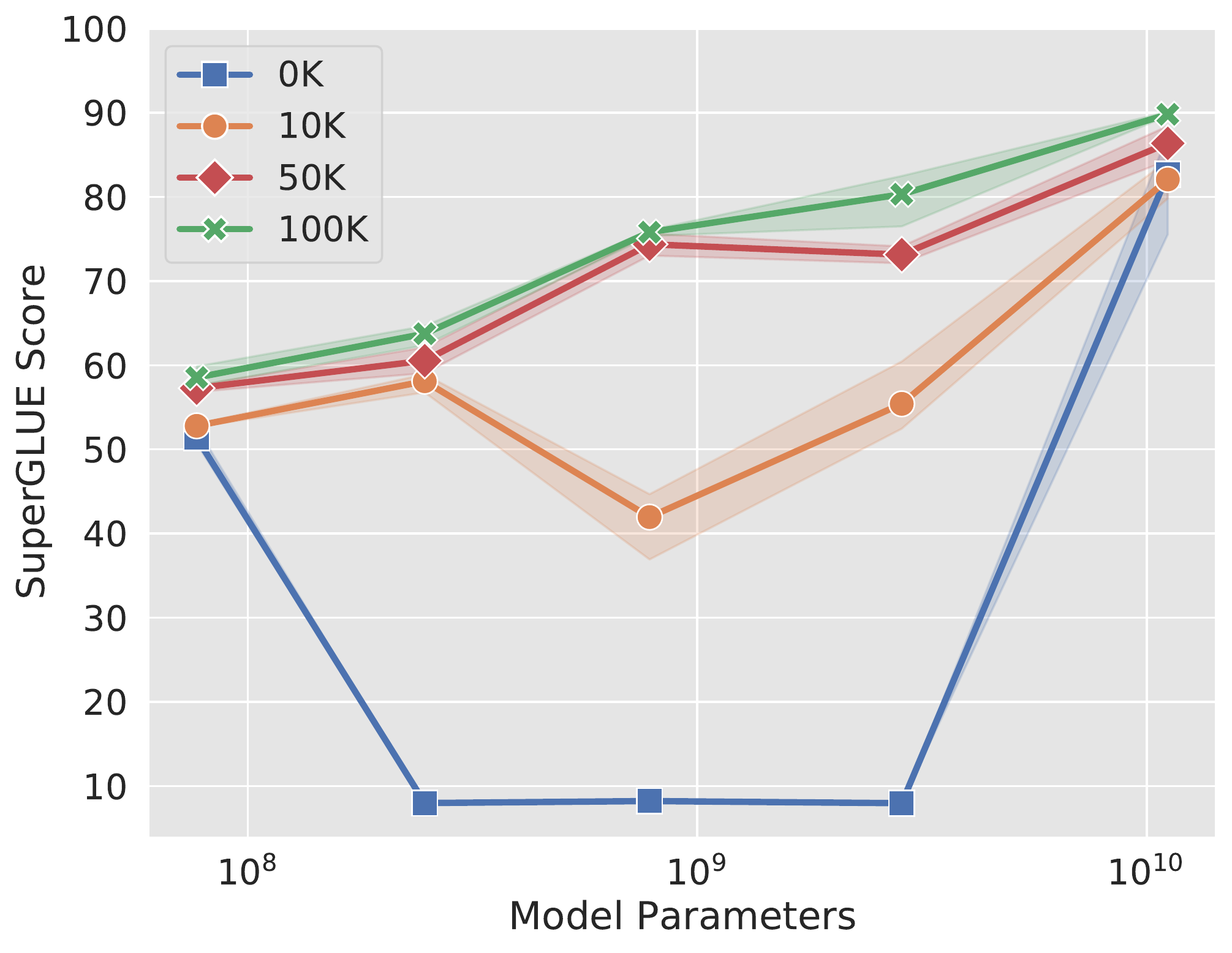}
     \caption{LM adaptation steps}
     \label{fig:ablate-lm-steps}
\end{subfigure}

\caption{Ablations of various hyperparameters on prompt tuning performance (mean and stddev across $3$ runs). In our ``\hyperref[settings:default]{default}'' (\inlinegraphics{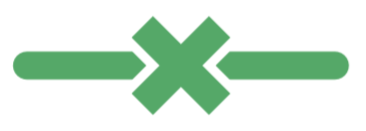}) configuration, quality improves stably with model size. Across all ablations, \emph{the largest (XXL) model is the most robust to hyperparameter choice}.
\subref{fig:ablate-length}~\textbf{Prompt length}: Increasing to $20$+ tokens generally confers a large boost, but XXL performs well even with single-token prompts.
\subref{fig:ablate-init}~\textbf{Prompt initialization}: Random uniform initialization lags behind more ``advanced'' initializations using sampled vocabulary or class label embeddings, but the difference vanishes at XXL size.
\subref{fig:ablate-pretrain}~\textbf{Pre-training objective}: LM adaptation outperforms span corruption, even when a sentinel is added to downstream task targets, but XXL works well with any method.
\subref{fig:ablate-lm-steps}~\textbf{LM adaptation}: Longer adaptation generally gives larger gains, but XXL is robust to even short adaptation.}
\label{fig:full-ablation}
\end{figure*}

\section{Results}
\label{sec:results}
Our frozen models are built on top of pre-trained T5 checkpoints of all sizes (Small, Base, Large, XL, XXL). We leverage the public T5.1.1 checkpoints, which include improvements over the original T5.\footnote{These improvements are (1) the removal of all supervised data from pre-training, (2) adjustments to hyperparameters $d_{\text{model}}$ and $d_{f\!f}$, and (3) the use of GeGLU \cite{shazeer2020glu} over ReLU \cite{Nair2010RectifiedLU} activations.}

\label{settings:default}Our ``default'' configuration, plotted with a green `$\times$' (\inlinegraphics{figures/green-x.png}) throughout, uses an LM-adapted version of T5 trained for an additional 100K steps, initializes using class labels (see Section~\ref{sec:ablations}), and uses a prompt length of $100$ tokens. While this is longer than the default 10-token prefix used by \citet{li_2021_prefix_tuning}, our method still uses fewer task-specific parameters, as we only tune the input layer, as opposed to overwriting activations in all network layers. See Figure~\ref{fig:param_counts} for a detailed comparison. We will also see shortly that even much shorter prompts are viable as model size increases.

We measure performance on the SuperGLUE benchmark \cite{wang2019superglue}, a collection of eight challenging English language understanding tasks.\footnote{The tasks are
BoolQ \cite{clark-etal-2019-boolq},
CB \cite{de-marneff_simons_tonhauser_2019},
COPA \cite{roemmele2011choice},
MultiRC \cite{MultiRC2018},
ReCoRD \cite{zhang2018record},
RTE \cite{dagan2005pascal,bar2006second,giampiccolo2007third,bentivogli2009fifth},
WiC \cite{DBLP:journals/corr/abs-1808-09121}, and
WSC \cite{levesque2012winograd}.} We report metrics on the development set associated with each dataset.

Each of our prompts train on a single SuperGLUE task; there was no multi-task setup or mixing of training data across tasks. We translate each SuperGLUE dataset into a text-to-text format following \citet{raffel_2020_t5}, except that we omit the task names prepended to inputs indicating which SuperGLUE task an example belongs to.

We train our prompts for $30{,}000$ steps using T5's standard cross-entropy loss, with a constant learning rate of $0.3$ and a batch size of $32$. Checkpoints are selected via early stopping on the development set, where the stopping metric is the default metric for the dataset, or the average of metrics for datasets evaluated with multiple metrics. All experiments were run in JAX \cite{jax2018github} using the Adafactor optimizer \cite{pmlr-v80-shazeer18a} with weight decay $1e{-5}$, $\beta_2$ decay $0.8$, and parameter scaling off. The models were implemented in Flax \cite{flax2020github}. More details are available in Appendix~\ref{app:reproducibility}.

\subsection{Closing the Gap}

To compare our method with standard model tuning, we tune the public T5.1.1 checkpoints on SuperGLUE using the default hyperparameters specified in the T5 library (learning rate $0.001$, and Adafactor optimizer with pre-training parameter states restored). We consider two baselines. (1)~``Model Tuning'': For an apples-to-apples comparison, we tune on each task separately, as in our prompt tuning setup.\footnote{To improve this baseline, we performed a sweep over the batch size hyperparameter and selected $2^{16}$ tokens per batch.} (2)~``Model Tuning (Multi-task)'': We use T5's multi-task tuning setup to achieve a more competitive baseline.\footnote{The T5 SuperGLUE submission used a more complex setup, first mixing multi-task supervised data into pre-training, and then performing single-task fine-tuning. Since we use T5.1.1 throughout, this setup is unavailable, as the pre-training phase is fully self-supervised. We follow \citet{raffel_2020_t5} in using $2^{20}$ tokens per batch and including DPR data in the multi-task mixture, which is known to boost WSC task performance \cite{kocijan-etal-2019-surprisingly}.} In this case, a single model is tuned on all tasks jointly, with a text prefix indicating the task name.

In Figure~\ref{fig:model-size} (p.~1), we see that prompt tuning becomes more competitive with model tuning as scale increases. At the XXL size (11 billion parameters), prompt tuning matches even the stronger multi-task model tuning baseline, despite having over $20{,}000$ times fewer task-specific parameters.

To compare with prompt design, we include GPT-3 few-shot performance on the SuperGLUE dev split, as reported by \citet{brown_2020_gpt3}.\footnote{We also experimented with using GPT-3's manual text prompts directly with our LM-adapted T5 checkpoints. However performance was far below GPT-3 for comparable model sizes. This may be due to differences in pre-training data and model architecture, as well as T5's shorter sequence length.} Figure~\ref{fig:model-size} shows that prompt tuning beats \mbox{GPT-3} prompt design by a large margin, with prompt-tuned \mbox{T5-Small} matching \mbox{GPT-3} XL (over $16$ times larger), and prompt-tuned \mbox{T5-Large} beating \mbox{GPT-3} 175B (over $220$ times larger).

\subsection{Ablation Study}
\label{sec:ablations}

\paragraph{Prompt Length} We train prompts for each model size while varying the prompt length in $\{1, 5, 20, 100, 150\}$ and fixing other settings to our \hyperref[settings:default]{default configuration}. Figure~\ref{fig:ablate-length} shows that for most model sizes, increasing prompt length beyond a single token is critical to achieve good performance. Notably, the XXL model still gives strong results with a single-token prompt, suggesting that the larger the model, the less conditioning signal is needed to achieve a target behavior. Across all models, increasing beyond $20$ tokens only yields marginal gains.\footnote{Going past $100$ tokens appears mildly detrimental for larger models. A similar pattern of diminishing performance past a certain prefix length is observed by \citet{li_2021_prefix_tuning}.}

\paragraph{Prompt Initialization} We ablate the effect of prompt initialization by training models at all sizes while fixing other hyperparameters to their \hyperref[settings:default]{default values}. For random initialization, we sample uniformly from the range [$-0.5$, $0.5$]. When initializing from sampled vocabulary, we restrict to the $5{,}000$ most ``common'' tokens in T5's SentencePiece vocabulary \cite{kudo2018sentencepiece}, which is ordered by likelihood in the pre-training corpus. For ``class label'' initialization, we take the embeddings for the string representations of each class in the downstream task and use them to initialize one of the tokens in the prompt. When a class label is multi-token, we average the token embeddings. At longer prompt lengths, we often run out of class labels before we have initialized all of the prompt tokens. In this case we fall back to our sampled vocab strategy to fill in the prompt.\footnote{T5's handling of the ReCoRD and WSC tasks requires the model to generate short, free-form text. In these cases, we initialize the prompts with words related to the task: \textit{commonsense}, \textit{reasoning}, \textit{reading}, and \textit{comprehension} for ReCoRD and \textit{commonsense}, \textit{pronoun}, and \textit{resolution} for WSC.}

Figure~\ref{fig:ablate-init} shows our ablation of initialization strategy across model sizes, where we find that the class based initialization performs best. At smaller model sizes, there are large gaps between the different initializations, but once the model is scaled to XXL size, those differences disappear.

With ``class label'' initialization, we observe that the class labels typically persist in the learned prompts, such that the nearest token embeddings (in cosine distance) match the tokens used for initialization. Beyond this, we did not find our learned prompts to be interpretable, similar to those of \citet{shin-etal-2020-autoprompt}. See Section~\ref{sec:interpretability} for details.

\paragraph{Pre-training Objective} In Figures~\ref{fig:ablate-pretrain} and \ref{fig:ablate-lm-steps}, we see pre-training objective has a clear effect on prompt tuning quality. As hypothesized in Section~\ref{sec:span_corruption}, T5's default ``span corruption'' objective is not well-suited for training frozen models to be later conditioned by prompts. Intuitively, models pre-trained to read and write sentinel tokens are hard to apply directly to tasks of reading and writing text without sentinels. As seen in Figure~\ref{fig:ablate-pretrain}, even the ``workaround'' of adding a sentinel to the downstream targets has little benefit. While LM adaptation adds value across all model sizes, we note our largest XXL model is the most forgiving and gives strong results even with span corruption.

Given the benefit of LM adaptation, we also explore how long of an adaptation is helpful. Figure~\ref{fig:ablate-lm-steps} shows that longer adaptation provides additional gains, up to $100$K steps. This suggests that the ``transition'' from span corruption to a language modeling objective is not a trivial change, and making an effective switch takes an investment of training resources ($10$\% of the steps of the original T5 pre-training). At the same time, as in our other ablations, we observe that the XXL model is robust to even non-ideal configurations. At this size, the gains from adaptation are quite modest.

In the non-optimal ``span corruption'' setting, we observe instability across model sizes, with the Small model outperforming the larger Base, Large, and XL models. On inspection, we find that for many tasks, these mid-sized models never learn to output a legal class label and thus score 0\%. The two most common error modes are copying sub-spans from the input and predicting an empty string. Furthermore, this poor performance is not due to random variance in prompt tuning, as we observe low variance across $3$ runs for each size. These results indicate that using models pre-trained with the ``span corruption'' objective can be unreliable, with only $2$ out of $5$ models working well, whereas the LM adapated versions work reliably across all model sizes.

We have released T5 1.1 checkpoints adapted using the LM objective for $100$K steps for all model sizes.\footnote{\url{https://github.com/google-research/text-to-text-transfer-transformer/blob/main/released_checkpoints.md##lm-adapted-t511lm100k}}

\section{Comparison to Similar Approaches}
\label{sec:previous_work}

\begin{figure}[t]
    \centering
    \includegraphics[width=0.9\columnwidth]{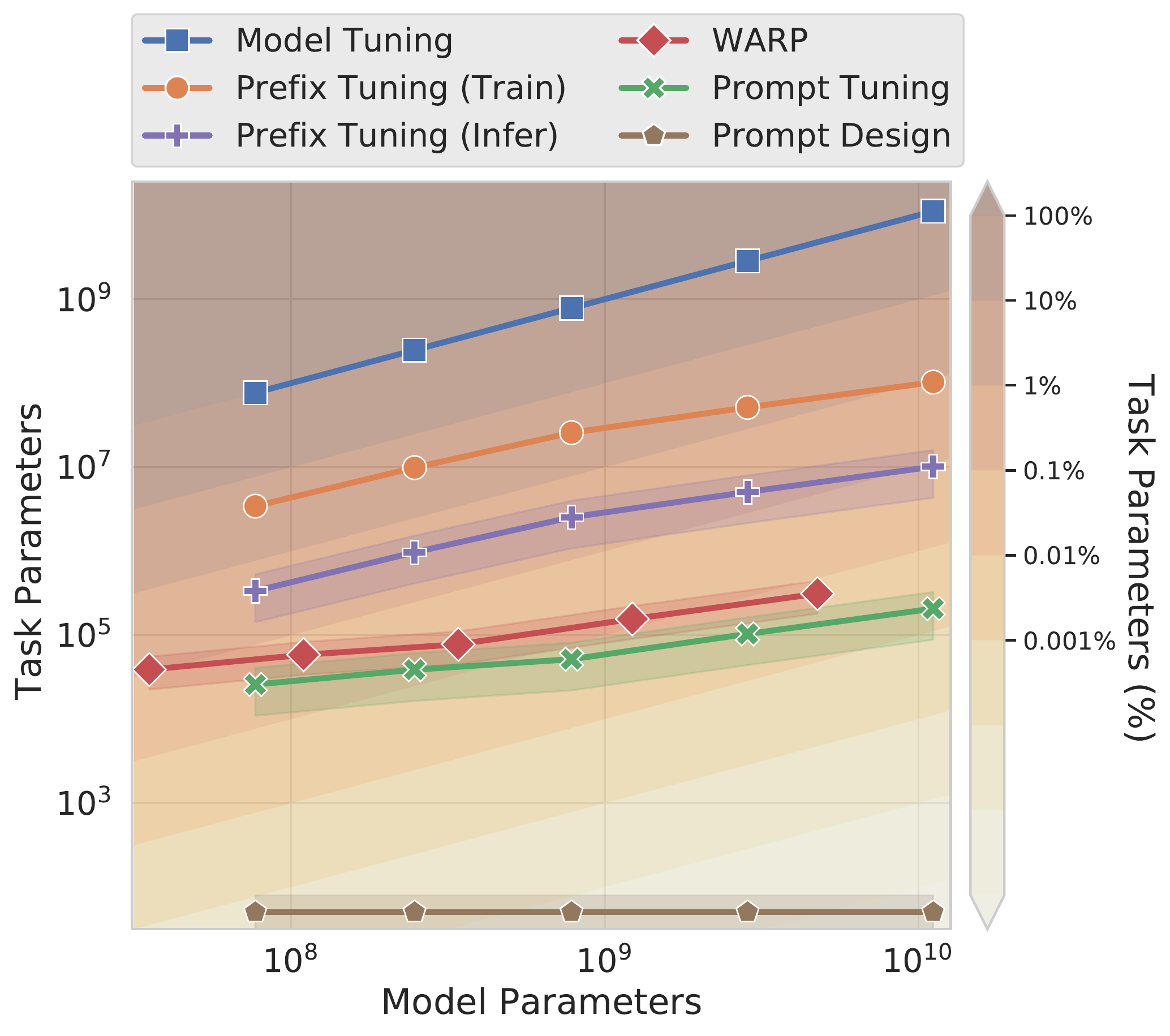}
    \caption{Parameter usage of various adaptation techniques, fixing architecture to T5.1.1 and prompt/prefix length to $1$--$100$ tokens (bands show mean and stddev). \textbf{Model Tuning}: All parameters are task-specific. \textbf{Prefix Tuning}: Activations are tuned in the prefix of each layer, requiring $0.1$--$1$\% task-specific parameters for inference, but more are used for training. \textbf{WARP}: Task parameters are reduced to under $0.1$\% by only tuning input and output layers. \textbf{Prompt Tuning}: Only prompt embeddings are tuned, reaching under $0.01$\% for most model sizes. \textbf{Prompt Design}: Only a sequence of prompt IDs ($500$--$2000$ tokens) is required.}
    \label{fig:param_counts}
\end{figure}

In this section, we review recent work on learning continuous prompts, and draw comparisons with our method. One important axis of comparison is the number of task-specific parameters each method requires, as shown in Figure~\ref{fig:param_counts}. Among methods with learnable parameters, prompt tuning is the most parameter efficient, requiring less than $0.01$\% task-specific parameters for models over a billion parameters.\footnote{To compare with prompt design, we count each token ID in the prompt as a parameter, and assume a prompt of between $500$--$2000$ tokens to match the GPT-3 setting. While this technique is by far the most parameter efficient, it comes at the cost of task quality.}

\citet{li_2021_prefix_tuning} propose ``prefix tuning'': learning a sequence of prefixes that are prepended at every transformer layer. This is akin to learning transformer activations that are fixed across examples at every network layer. In contrast, prompt tuning uses a single prompt representation that is prepended to the embedded input. Beyond requiring fewer parameters, our approach allows the transformer to update the intermediate-layer task representations, as contextualized by an input example. Their work builds on GPT-2 \cite{radford2019language} and BART \cite{lewis-etal-2020-bart}, while ours focuses on T5 and examines changes in performance and robustness to design choices as model size increases. When using BART, prefix tuning includes prefixes on both the encoder and decoder network, while prompt tuning only requires prompts on the encoder. \citet{li_2021_prefix_tuning} also rely on a reparameterization of the prefix to stabilize learning, which adds a large number of parameters during training, whereas our configuration does not require this reparameterization and is robust across SuperGLUE tasks and model sizes.

\citet{hambardzumyan_2021_warp} propose ``WARP'', where prompt parameters are added to the input layer. This method works with masked language models, relying on a {\tt [MASK]} token and a learnable output layer to project the mask to class logits. This formulation restricts the model to producing a single output, limiting it to classification. Prompt tuning does not require any changes to the input or a task-specific head. The performance of prompt tuning is also considerably closer to the strong performance of model tuning.

\citet{liu2021gpt} propose ``P-tuning'' where learnable continuous prompts are interleaved throughout the embedded input, using patterns based on human design. Our approach removes this complication by simply prepending the prompt to the input. To achieve strong SuperGLUE results, P-tuning has to be used in \emph{conjunction} with model tuning, that is, models jointly update both the prompt and the main model parameters, whereas our approach keeps the original language model frozen.\footnote{As another difference, P-tuning requires the addition of ``anchor'' tokens in the input (e.g.\ a question mark following the hypothesis in the RTE task) to achieve strong performance, while prompt tuning leaves inputs untouched.}

\citet{qinLearningHowAsk2021} use ``soft words'' to learn prompts to extract knowledge from pre-trained LMs. Prompts are positioned in relation to the input based on hand-designed prompt prototypes, and a learned $\Delta_i^{\ell}$ parameter is included for each layer, so parameter cost scales with model depth.

\citet{logeswaranFewshotSequenceLearning2020} use a learnable prepended token to adapt transformer models to various tasks, but focus on small synthetic datasets designed to accommodate a compositional task representation, as opposed to larger real-world datasets. Their base models are small transformers trained from scratch \emph{jointly} with the task representations, whereas we keep the base model frozen and investigate scaling laws using larger transformers.

More generally, work on task prompts is closely aligned with work on ``adapters'' \cite{rebuffi_2017_adapters, houlsby_2019_adapters}, small bottleneck layers inserted \emph{between} frozen pre-trained network layers. Adapters offer another means of reducing task-specific parameters, with \citet{houlsby_2019_adapters} achieving GLUE performance close to full model tuning when freezing BERT-Large and only adding $2$--$4$\% additional parameters. \citet{pfeiffer-etal-2020-mad} use multiple adapters in a multilingual context to explicitly separate language understanding from task specification, similar to our approach. A core difference between adapters and prompt tuning is how the approaches change model behavior. Adapters modify the actual function that acts on the input representation, parameterized by the neural network, by allowing the rewriting of activations at any given layer. Prompt tuning modifies behavior by leaving the function fixed and adding new input representations that can affect how subsequent input is processed.

\section{Resilience to Domain Shift}
\label{sec:shift}

\begin{table}[]
\centering
\footnotesize
\resizebox{0.9\columnwidth}{!}{
\begin{tabular}{ll|ccr}
\toprule
\textbf{Dataset} & \textbf{Domain} & \textbf{Model} & \textbf{Prompt} & $\Delta$ \\
\midrule
SQuAD & Wiki & 94.9 $\pm$0.2 & 94.8 $\pm$0.1 & $-$0.1 \\
\midrule
TextbookQA & Book & 54.3 $\pm$3.7 & \textbf{66.8} $\pm$2.9 & +12.5 \\
BioASQ & Bio & 77.9 $\pm$0.4 & \textbf{79.1} $\pm$0.3 & +1.2 \\
RACE & Exam & 59.8 $\pm$0.6 & \textbf{60.7} $\pm$0.5 & +0.9 \\
RE & Wiki & 88.4 $\pm$0.1 & \textbf{88.8} $\pm$0.2 & +0.4 \\
DuoRC & Movie & \textbf{68.9} $\pm$0.7  & 67.7 $\pm$1.1 & $-$1.2 \\
DROP & Wiki & \textbf{68.9} $\pm$1.7  & 67.1 $\pm$1.9 & $-$1.8 \\ 
\bottomrule
\end{tabular}
}
\caption{
F1 mean and stddev for models trained on SQuAD and evaluated on out-of-domain datasets from the MRQA 2019 shared task. Prompt tuning tends to give stronger zero-shot performance than model tuning, especially on datasets with large domain shifts like TextbookQA. 
}
\label{tab:domain-squad}
\end{table}

By freezing the core language model parameters, prompt tuning prevents the model from modifying its general understanding of language. Instead, prompt representations indirectly modulate the representation of the input. This reduces the model's ability to overfit to a dataset by memorizing specific lexical cues and spurious correlations. This restriction suggests that prompt tuning may improve robustness to domain shifts, where the distribution of inputs differs between training and evaluation.

We investigate zero-shot domain transfer on two tasks: question answering (QA) and paraphrase detection. For question answering, we use the MRQA 2019 shared task on generalization \cite{fisch2019mrqa}. This task collects extractive QA datasets in a unified format and tests how models trained on ``in-domain'' datasets perform when evaluated on ``out-of-domain'' datasets. For our experiments, we train on SQuAD \cite{rajpurkar-etal-2016-squad} and evaluate on each of the out-of-domain datasets.\footnote{We select checkpoints based on SQuAD validation F1. The out-of-domain datasets are TextbookQA \cite{8100054}, RACE \cite{lai-etal-2017-race}, BioASQ (\url{http://bioasq.org/}), RE \cite{levy-etal-2017-zero}, DuoRC \cite{saha-etal-2018-duorc}, and DROP \cite{dua-etal-2019-drop}.}

Table~\ref{tab:domain-squad} shows that prompt tuning outperforms model tuning on the majority of out-of-domain datasets, with a remarkable $12.5$ point F1 gap between the two approaches on TextbookQA. We observe larger gains from prompt tuning in cases of larger domain shifts (e.g.~to Biomedical in BioASQ or to Textbooks in TextbookQA). Of the datasets where model tuning is better, we see that DROP shares a domain (Wikipedia) with SQuAD and is thus one of the smallest domain transfers.

As a second test of robustness to domain shift, we explore transfer between two paraphrase detection tasks from GLUE \cite{wang2019glue}. The first task is QQP \cite{WinNT}, which asks if two questions from the community Q\&A site Quora are ``duplicates''. The second task is MRPC \cite{dolan2005automatically}, which asks if two sentences drawn from news articles are paraphrases. We test transfer in both directions (QQP $\Leftrightarrow$ MRPC). As before, we train on the ``in-domain'' task, select checkpoints using in-domain validation, and evaluate zero-shot on the ``out-of-domain'' task.

\begin{table}[]
\footnotesize
\centering
\resizebox{0.85\columnwidth}{!}{
\begin{tabular}{lll|cc}
    \toprule
    \textbf{Train} & \textbf{Eval} & \textbf{Tuning} & \textbf{Accuracy} & \textbf{F1}  \\
    \midrule
 QQP  & MRPC & Model  & 73.1 $\pm$0.9 & 81.2 $\pm$2.1 \\ 
      &      & Prompt & \textbf{76.3} $\pm$0.1 & \textbf{84.3} $\pm$0.3 \\
      \midrule
 MRPC & QQP  & Model  & 74.9 $\pm$1.3  & \textbf{70.9} $\pm$1.2  \\
      &      & Prompt & \textbf{75.4} $\pm$0.8 & 69.7 $\pm$0.3 \\
    \bottomrule
\end{tabular}
}
\caption{
    Mean and stddev of zero-shot domain transfer between two paraphrase detection tasks.
}
\label{tab:domain-paraphrase}
\end{table}

Table~\ref{tab:domain-paraphrase} shows that training a lightweight prompt on the QQP data and evaluating on MRPC gives much better performance than tuning the entire model (+$3.2$ accuracy and +$3.1$ F1). The results are much closer in the other direction, with prompt tuning showing a small improvement in accuracy and a small drop in F1\@. These results support the view that model tuning may be over-parameterized and more prone to overfit the training task, to the detriment of similar tasks in different domains.

\section{Prompt Ensembling}
\label{sec:ensemble}

Ensembles of neural models trained from different initializations on the same data are widely observed to improve task performance \cite{hansen_1990_ensembles} and are useful for estimating model uncertainty \cite{lakshminarayanan_2017_deep_ensembles}. However, as model size increases, ensembling can become impractical. Beyond the space required to store $N$ models (e.g.\ $42$ GiB for each copy of T5-XXL), there is a substantial inference cost to running $N$ distinct models, whether in parallel or in series.

Prompt tuning provides a more efficient way to ensemble multiple adaptations of a pre-trained language model. By training $N$ prompts on the same task, we create $N$ separate ``models'' for a task, while still sharing the core language modeling parameters throughout. Beyond drastically reducing storage costs, the prompt ensemble makes inference more efficient. To process one example, rather than computing forward passes of $N$ different models, we can execute a single forward pass with a batch size of $N$, replicating the example across the batch and varying the prompt. These savings mirror those seen for multi-tasking in Figure~\ref{fig:diagram}.

\begin{table}[t]
\setlength\tabcolsep{5pt}
\centering
\footnotesize
\resizebox{\columnwidth}{!}{
\begin{tabular}{ll|ccc}
\toprule
\textbf{Dataset} & \textbf{Metric} & \textbf{Average} & \textbf{Best} & \textbf{Ensemble}  \\
\midrule
BoolQ   & acc. & 91.1 & 91.3 &  \textbf{91.7} \\
CB      & acc./F1 & 99.3 / 99.0 & 100.00 / 100.00 & \textbf{100.0} / \textbf{100.0} \\
COPA    & acc. & 98.8 & 100.0 &  \textbf{100.0} \\
MultiRC & EM/F1$_a$ & 65.7 / 88.7 & 66.3 / 89.0 &  \textbf{67.1} / \textbf{89.4} \\
ReCoRD  & EM/F1  & 92.7 / 93.4 & 92.9 / 93.5 &  \textbf{93.2} / \textbf{93.9} \\
RTE     & acc. & 92.6 & \textbf{93.5} &  \textbf{93.5} \\
WiC     & acc. & 76.2 & 76.6 &  \textbf{77.4} \\
WSC     & acc. & 95.8 & \textbf{96.2} &  \textbf{96.2} \\
\midrule
\multicolumn{2}{l}{SuperGLUE (dev)} & 90.5 & 91.0 &  \textbf{91.3} \\
\bottomrule
\end{tabular}
}
\caption{
Performance of a five-prompt ensemble built from a single frozen T5-XXL model exceeds both the average and the best among the five prompts.}
\label{tab:ensemble}
\end{table}

To demonstrate the viability of prompt ensembling, we train five prompts for each SuperGLUE task, using a single frozen T5-XXL model with our default hyperparameters. We use simple majority voting to compute predictions from the ensemble. Table~\ref{tab:ensemble} shows that across all tasks, the ensemble beats the single-prompt average and beats, or matches, the best individual prompt.

\section{Interpretability}
\label{sec:interpretability}

An ideally interpretable prompt would consist of natural language that clearly describes the task at hand, explicitly asks the model for some result or action, and makes it easy to understand why the prompt elicited such behavior from the model.

As prompt tuning works in the continuous embedding space rather than the discrete token space, interpreting prompts becomes more difficult. To test the interpretability of our learned soft prompts, we compute the nearest neighbors to each prompt token from the frozen model's vocabulary. We use cosine distance between the vocabulary embedding vector and the prompt token representation as the similarity metric.

We observe that for a given learned prompt token, the top-5 nearest neighbors form tight semantic clusters. For example, we see lexically similar clusters such as \{~\textit{Technology} / \textit{technology} / \textit{Technologies} / \textit{technological} / \textit{technologies}~\}, as well as more diverse but still strongly related clusters such as \{~\textit{entirely} / \textit{completely} / \textit{totally} / \textit{altogether} / \textit{100\%}~\}. The nature of these clusters suggests that the prompts are in fact learning ``word-like'' representations. We found that random vectors drawn from the embedding space do not show this sort of semantic clustering.

When initializing the prompts using the ``class-label'' strategy, we often find that the class labels persist through training. Specifically, if a prompt token is initialized to a given label, that label is often among the learned token's nearest neighbors after tuning. When initializing with the ``Random Uniform'' or ``Sampled Vocab'' methods, the class labels can also be found in the nearest neighbors of the prompts; however they tend to appear as neighbors to multiple prompt tokens. This suggests that the model is learning to store the expected output classes in the prompts as reference, and initializing the prompt to outputs classes makes this easier and more centralized.

When examining longer prompts (e.g.~size $100$), we often find several prompt tokens with the same nearest neighbors. This suggests there is either excess capacity in the prompt, or that the lack of sequential structure in the prompt representation makes it difficult for the model to localize information to a specific position.

While the learned prompts taken as sequences show little interpretability, we do observe a high frequency of words like \textit{science}, \textit{technology} and \textit{engineering} as the nearest neighbors for prompts trained on the BoolQ dataset and approximately $20$\% of the questions are in the ``Nature/Science'' category. While more investigation is needed, this suggests that one role of the prompt may be to prime the model to interpret inputs in a specific domain or context (e.g.~``scientific'').

\section{Conclusion}

In this paper, we showed that prompt tuning is a competitive technique for adapting frozen pre-trained language models to downstream tasks. On the popular SuperGLUE benchmark, its task performance rivals that of traditional model tuning, with the gap vanishing as model size increases. On zero-shot domain transfer, we found that prompt tuning leads to improved generalization. This plausibly indicates that freezing general-purpose language understanding parameters and restricting downstream learning to a lightweight parameter footprint can help to avoid overfitting to a specific domain.

Beyond task quality metrics, we discussed the appeal of moving to frozen pre-trained models in terms of storage and serving costs. This move enables both efficient multi-task serving, as well as efficient high-performing prompt ensembling. Looking forward, we believe that factoring out task-defining parameters as distinct from general language-modeling parameters is an exciting step that opens up many avenues for new research.

\section*{Acknowledgements}

We thank Lucas Dixon, Waleed Ammar, Slav Petrov and Sebastian Ruder for comments on an earlier draft, and the following people for helpful discussion: Colin Raffel, Adam Roberts, and Noam Shazeer. We thank Linting Xue for help with the LM adaptation training.

\bibliography{anthology,custom}
\bibliographystyle{acl_natbib}

\clearpage

\appendix

\section{Reproducibility}
\label{app:reproducibility}

\subsection{Experimental Settings}

We evaluate each GLUE and SuperGLUE dataset using the metric specified in the benchmark. We reuse the evaluation code from the publicly available T5 open-source release to compute metrics.\footnote{\url{https://github.com/google-research/text-to-text-transfer-transformer/blob/master/t5/evaluation/metrics.py}} For the SQuAD and MRQA datasets, we evaluate using F1, one of the metrics used by the SQuAD benchmark, where partial answer spans are considered. Again, we use the T5 open-source release for metric calculation.\footnote{\url{https://github.com/google-research/text-to-text-transfer-transformer/blob/master/t5/evaluation/metrics.py##L151}} All of our models use T5 1.1 as the base frozen model, additional details and pre-trained checkpoints can be found on GitHub.\footnote{\url{https://github.com/google-research/text-to-text-transfer-transformer/blob/master/released_checkpoints.md##t511}}\footnote{\url{https://github.com/google-research/text-to-text-transfer-transformer/blob/main/released_checkpoints.md##lm-adapted-t511lm100k}}  

All prompts for T5 Small and Base models were trained on $4$ TPU v2 chips, while prompts for larger models were trained on $16$ TPU v3 chips.

\begin{table*}[t]
    \centering
    \footnotesize
    \begin{tabular}{l r |r r r}
    \toprule
    T5 Size & Prompt Length & Trainable Parameters & Total Parameters & Percent Trainable \\
    \midrule
    Small   & $1$           & $512$                & $76{,}961{,}664$       &  $0.00067$\% \\
            & $5$           & $2{,}560$            & $76{,}963{,}712$       &  $0.00333$\% \\
            & $20$          & $10{,}420$           & $76{,}971{,}572$       &  $0.01330$\% \\
            & $50$          & $25{,}600$           & $76{,}986{,}752$       &  $0.03325$\% \\
            & $100$         & $51{,}200$           & $77{,}012{,}352$       &  $0.06648$\% \\
            & $150$         & $76{,}800$           & $77{,}037{,}952$       &  $0.09969$\% \\
    Base    & $1$           & $768$                & $247{,}578{,}624$      &  $0.00031$\% \\
            & $5$           & $3{,}840$            & $247{,}581{,}696$      &  $0.00155$\% \\
            & $20$          & $15{,}360$           & $247{,}593{,}216$      &  $0.00620$\% \\
            & $50$          & $38{,}400$           & $247{,}616{,}256$      &  $0.01551$\% \\
            & $100$         & $76{,}800$           & $247{,}654{,}656$      &  $0.03101$\% \\
            & $150$         & $115{,}200$          & $247{,}693{,}056$      &  $0.04651$\% \\
    Large   & $1$           & $1{,}024$            & $783{,}151{,}104$      &  $0.00013$\% \\
            & $5$           & $5{,}120$            & $783{,}155{,}200$      &  $0.00065$\% \\
            & $20$          & $20{,}480$           & $783{,}170{,}560$      &  $0.00262$\% \\
            & $50$          & $51{,}200$           & $783{,}201{,}280$      &  $0.00654$\% \\
            & $100$         & $102{,}400$          & $783{,}252{,}480$      &  $0.01907$\% \\
            & $150$         & $153{,}600$          & $783{,}303{,}680$      &  $0.01961$\% \\
    XL      & $1$           & $2{,}048$            & $2{,}849{,}759{,}232$  &  $0.00007$\% \\
            & $5$           & $10{,}240$           & $2{,}849{,}767{,}424$  &  $0.00036$\% \\
            & $20$          & $40{,}960$           & $2{,}849{,}798{,}144$  &  $0.00143$\% \\
            & $50$          & $102{,}400$          & $2{,}849{,}859{,}584$  &  $0.00359$\% \\
            & $100$         & $204{,}800$          & $2{,}849{,}961{,}984$  &  $0.00718$\% \\
            & $150$         & $307{,}200$          & $2{,}850{,}064{,}384$  &  $0.01078$\% \\
    XXL     & $1$           & $4{,}096$            & $11{,}135{,}336{,}448$ &  $0.00004$\% \\
            & $5$           & $20{,}480$           & $11{,}135{,}352{,}832$ &  $0.00018$\% \\
            & $20$          & $81{,}920$           & $11{,}135{,}414{,}272$ &  $0.00074$\% \\
            & $50$          & $204{,}800$          & $11{,}137{,}380{,}352$ &  $0.00184$\% \\
            & $100$         & $409{,}600$          & $11{,}135{,}741{,}952$ &  $0.00368$\% \\
            & $150$         & $614{,}400$          & $11{,}135{,}946{,}752$ &  $0.00552$\% \\
    \bottomrule
    \end{tabular}
    \caption{Number of parameters used for various prompt lengths and T5 model sizes. Trainable parameters is the number of parameters in the prompt itself, while total parameters includes the prompt plus the original T5 parameters. The T5 parameters are frozen and shared across all tasks, and include the SentencePiece lookup table parameters. The final column is the percentage of total parameters that are trainable.}
    \label{tab:prompt-parameter-counts}
\end{table*}

Parameter counts for each prompt can be found in Table~\ref{tab:prompt-parameter-counts}. Average runtimes until convergence can be found in Table~\ref{tab:runtime}.

\begin{table}[t]
    \centering
    \footnotesize
    \begin{tabular}{l l| r}
    \toprule
    Prompt Length & T5 Size & Time \\
    \midrule
        1   & Large &    3:17 $\pm$02:10 \\
            & XL    &    3:37 $\pm$02:11 \\
            & XXL   &   21:23 $\pm$01:54 \\
        20  & XL    &   49:08 $\pm$18:53 \\
            & XXL   &   53:03 $\pm$16:25 \\
        50  & Small &   09:05 $\pm$05:07 \\
            & Base  &   55:01 $\pm$27:48 \\
            & Large & 1:14:16 $\pm$13:12 \\
            &  XL   & 2:30:10 $\pm$25:40 \\
            & XXL   & 3:13:13 $\pm$23:08 \\
        100 & Small &   16:25 $\pm$01:15 \\
            & Base  &   29:57 $\pm$00:18 \\        
            & Large & 1:23:36 $\pm$10:21 \\
            &  XL   & 3:35:00 $\pm$54:42 \\
            & XXL   & 3:51:15 $\pm$45:53 \\
    \bottomrule
    \end{tabular}
    \caption{Mean and standard deviation of the runtime until convergence for the BoolQ dataset and various prompt lengths and model sizes. Convergence is defined as reaching a performance within $1$\% of the mean value for that model configuration. A few configurations have been omitted because their runtimes were artificially extended due to preemption.}
    \label{tab:runtime}
\end{table}


\subsection{Hyperparameter Search}

This work used $77$ hyperparameter search trials ($40$ for prompt tuning and $37$ for single-task model tuning), and $3$ training runs (with validation evaluation) for each baseline configuration and ablation setting, for a total of $195$ runs for our main result and ablations. There were an additional $18$ runs for the domain shift experiments and $24$ extra runs to create the ensemble. Hyperparameter bounds can be found in Table~\ref{tab:hyperparam-bounds}. Hyperparameter tuning was done via manual tuning and settings were selected based on the SuperGLUE score. All experiments in this work, outside of the hyperparameter being ablated, use our \hyperref[settings:default]{default configuration} of 100K steps of LM Adapation, a prompt length of 100, and ``class-label'' initialization.

All graphs of our experimental results plot the mean and standard deviation over 3 runs as computed by Seaborn \cite{Waskom2021}. Some settings have such low variance that the standard deviation is hidden behind the line itself, such as ``Model Tuning (Multi-task)'' in Figure~\ref{fig:model-size} and the Base, Large, and XL prompts trained on the ``Span Corruption'' pretraining objective in Figure~\ref{fig:ablate-init}. Figure~\ref{fig:param_counts} also shows mean and standard deviation for the number of parameters each method uses as the prompt length varies from $1$--$100$. The ``Prefix Tuning (Train)'' curves appears to have no standard deviation because the parameter count is so strongly dominated by the cost of the reparameterization parameters that the standard deviation bands are occluded. For our experiments on domain transfer, we report mean and standard deviation over 3 runs.

\begin{table}[t]
    \centering
    \footnotesize
    \begin{tabular}{l|r}
    \toprule
    Hyperparameter & Search Space \\
    \midrule
    Learning Rate  & $0.001$--$0.5$  \\
    Parameter Scaling & $\{$True, False$\}$ \\
    Batch Size & $\{32, 64, 126, 256, 512\}$ \\
    Number of Steps & $\{10{,}000, 20{,}000, 30{,}000\}$ \\
    Warmup Steps & $\{\text{off}, 2{,}000, 3{,}000 \}$ \\
    Decay Factor & $\{\text{off}, 0.1, 0.5\}$ \\
    Steps per Decay & $\{\text{off}, 4{,}000, 6{,}000, 8{,}000 \}$ \\
    \bottomrule
    \end{tabular}
    \caption{Search space for each hyperparameter considered. Parameter Scaling refers to the Adafactor setting where an update is scaled by the norm of the parameter it will be applied to. Warmup Steps is the number of steps before a linearly increasing learning rate reaches the Learning Rate value, starting from zero. Decay Factor is the reduction in Learning Rate size that occurs every ``Steps per Decay'' steps.}
    \label{tab:hyperparam-bounds}
\end{table}

\subsection{Datasets}

All datasets used are in English. For the GLUE\footnote{\url{https://www.tensorflow.org/datasets/catalog/glue##gluemrpc}}\textsuperscript{,}\footnote{\url{https://www.tensorflow.org/datasets/catalog/glue##glueqqp}} and SuperGLUE\footnote{\url{https://www.tensorflow.org/datasets/catalog/super_glue}} datasets, we used the training, validation, and test splits that ship with TensorFlow Datasets. We used version {\tt 1.0.0} for GLUE and {\tt 1.0.2} for SuperGLUE datasets. For SQuAD\footnote{\url{https://www.tensorflow.org/datasets/catalog/squad##squadv11_default_config}} we used {\tt v1.1:3.0.0} from Tensorflow Datasets and follow the provided training, validation, and test splits. For the out-of-domain datasets we used the development splits distributed as part of the MRQA shared task.\footnote{\url{https://github.com/mrqa/MRQA-Shared-Task-2019##out-of-domain}} Dataset sizes can be found in Table~\ref{tab:dataset-size}. The label distributions for each dataset can be found in Table~\ref{tab:boolq-label-dist} (BoolQ), Table~\ref{tab:cb-label-dist} (CB), Table~\ref{tab:copa-label-dist} (COPA), Table~\ref{tab:multirc-label-dist} (MultiRC), Table~\ref{tab:rte-label-dist} (RTE), Table~\ref{tab:wic-label-dist} (WiC), Table~\ref{tab:wsc-label-dist} (WSC), Table~\ref{tab:mrpc-label-dist} (MRPC) and Table~\ref{tab:qqp-label-dist} (QQP).

\begin{table}[t]
    \centering
    \footnotesize
    \begin{tabular}{l| r r r}
        \toprule
        Dataset    & Training    & Validation & Testing     \\
        \midrule
        BoolQ      & $9{,}427$   & $3{,}270$  & $3{,}245$   \\
        CB         & $250$       & $56$       & $250$       \\
        COPA       & $400$       & $100$      & $500$       \\
        MultiRC    & $27{,}243$  & $4{,}848$  & $9{,}693$   \\
        ReCoRD     & $100{,}730$ & $10{,}000$ & $10{,}000$  \\
        RTE        & $2{,}490$   & $277$      & $3{,}000 $  \\
        WiC        & $5{,}428$   & $638$      & $1{,}400$   \\
        WSC        & $259^*$     & $104$      & $146$       \\
        MRPC       & $3{,}668$   & $408$      & $1{,}725$   \\
        QQP        & $363{,}849$ & $40{,}430$ & $390{,}965$ \\
        SQuAD      & $87{,}599$  & $10{,}570$ &           - \\
        TextbookQA &           - & $1{,}504$  &           - \\
        RACE       &           - & $1{,}503$  &           - \\
        BioASQ     &           - & $1{,}501$  &           - \\
        RE         &           - & $674$      &           - \\
        DuoRC      &           - & $2{,}948$  &           - \\
        DROP       &           - & $1{,}503$  &           - \\
        \bottomrule
    \end{tabular}
    \caption{Sizes for training, validation, and testing splits of each dataset used. $^*$Following T5, our casting of WSC as a text generation problems means we can only train on examples where the supplied referent is correct. This means our training dataset is smaller than the normal WSC training dataset, which has $554$ examples.}
    \label{tab:dataset-size}
\end{table}

\begin{table}[t]
    \centering
    \footnotesize
    \begin{tabular}{l|r r}
    \toprule
    Split & {\tt False} & {\tt True} \\
    \midrule
    Training & $37.7$ & $62.3$ \\
    Validation & $37.8$ & $62.2$ \\
    \bottomrule
    \end{tabular}
    \caption{Label distribution for the BoolQ dataset.}
    \label{tab:boolq-label-dist}
\end{table}

\begin{table}[t]
    \centering
    \footnotesize
    \resizebox{\columnwidth}{!}{
    \begin{tabular}{l| r r r}
    \toprule
    Split & {\tt contradiction} & {\tt entailment} & {\tt neutral} \\
    \midrule
    Training & $47.6$ & $46.0$ & $6.4$  \\
    Validation & $50.0$ & $41.1$ & $8.9$ \\
    \bottomrule
    \end{tabular}
    }
    \caption{Label distribution for the CB dataset.}
    \label{tab:cb-label-dist}
\end{table}

\begin{table}[t]
    \centering
    \footnotesize
    \begin{tabular}{l| r r}
    \toprule
    Split & {\tt choice1} & {\tt choice2} \\
    \midrule
    Training & $48.8$ & $51.2$ \\
    Validation & $55.0$ & $45.0$ \\
    \bottomrule
    \end{tabular}
    \caption{Label distribution for the COPA dataset.}
    \label{tab:copa-label-dist}
\end{table}

\begin{table}[t]
    \centering
    \footnotesize
    \begin{tabular}{l| r r}
    \toprule
    Split & {\tt False} & {\tt True} \\
    \midrule
    Training & $55.9$ & $44.1$ \\
    Validation & $57.2$ & $42.8$ \\
    \bottomrule
    \end{tabular}
    \caption{Label distribution for the MultiRC dataset.}
    \label{tab:multirc-label-dist}
\end{table}

\begin{table}[t]
    \centering
    \footnotesize
    \begin{tabular}{l | r r}
    \toprule
    Split & {\tt False} & {\tt True} \\
    \midrule
    Training & $50.0$ & $50.0$ \\
    Validation & $50.0$ & $50.0$ \\
    \bottomrule
    \end{tabular}
    \caption{Label distribution for the WiC dataset.}
    \label{tab:wic-label-dist}
\end{table}

\begin{table}[t]
    \centering
    \footnotesize
    \begin{tabular}{l | r r}
    \toprule
    Split & {\tt False} & {\tt True} \\
    \midrule
    Training & $0.0$ & $100.0$ \\
    Validation & $63.5$ & $36.5$ \\
    \bottomrule
    \end{tabular}
    \caption{Label distribution for the WSC dataset. Following T5, we cast the WSC dataset to a free-form text generation task where the model generates the referent to the highlighted span instead predicting if the supplied entity is the correct referent of the highlighted span. Thus, we only use training data where the supplied referent is correct making our training label distribution focused entirely on {\tt True}.}
    \label{tab:wsc-label-dist}
\end{table}

\begin{table}[t]
    \centering
    \footnotesize
    \begin{tabular}{l | r r}
    \toprule
    Split & {\small\tt entailment} & {\small\tt not\_entailment} \\
    \midrule
    Training & $51.2$ & $49.8$ \\
    Validation & $52.7$ & $47.3$ \\
    \bottomrule
    \end{tabular}
    \caption{Label distribution for the RTE dataset.}
    \label{tab:rte-label-dist}
\end{table}

\begin{table}[t]
    \centering
    \footnotesize
    \begin{tabular}{l | r r}
    \toprule
    Split & {\small\tt equivalent} & {\small\tt not\_equivalent} \\
    \midrule
    Training & $67.4$ & $32.6$ \\
    Validation & $68.4$ & $31.6$ \\
    \bottomrule
    \end{tabular}
    \caption{Label distribution for the MRPC dataset.}
    \label{tab:mrpc-label-dist}
\end{table}

\begin{table}[t]
    \centering
    \footnotesize
    \begin{tabular}{l | r r}
    \toprule
    Split & {\small\tt duplicate} & {\small\tt not\_duplicate} \\
    \midrule
    Training & $36.9$ & $63.1$ \\
    Validation & $36.8$ & $63.2$ \\
    \bottomrule
    \end{tabular}
    \caption{Label distribution for the QQP dataset.}
    \label{tab:qqp-label-dist}
\end{table}

The question answering datasets are extractive datasets with a variety of answers, so there isn't a label distribution to report. Similarly, the ReCoRD dataset is a multiple choice dataset where the model must predict the masked out entity from a list of possible entities. Due to this formulation there isn't a meaningful label distribution.

We followed the open-source T5 preprocessing procedure{\interfootnotelinepenalty=10000 \footnote{\url{https://github.com/google-research/text-to-text-transfer-transformer/blob/master/t5/data/preprocessors.py}}} for each dataset, except that we omit the dataset prefix denoting which SuperGLUE dataset an example belongs to. For the SQuAD and MRQA datasets we used the T5 SQuAD preprocessing code{\interfootnotelinepenalty=10000\footnote{\url{https://github.com/google-research/text-to-text-transfer-transformer/blob/master/t5/data/preprocessors.py##L264}}}. By following the T5 preprocessing and text-to-text format, we recast the WSC dataset as a text generation task. Instead of predicting whether a supplied referent is correct for a highlighted span, our model predicts the correct referent directly. As such, we can only learn from training examples where the referent is correct, so WSC training data where the supplied referent is incorrect are omitted.

No new data was collected for this work.

\end{document}